\documentclass{article}

\usepackage[numbers]{natbib}
\usepackage[preprint]{neurips_2025}

\usepackage[utf8]{inputenc} 
\usepackage[T1]{fontenc}    
\usepackage{hyperref}       
\usepackage{url}            
\usepackage{booktabs}       
\usepackage{amsfonts}       
\usepackage{nicefrac}       
\usepackage{microtype}      
\usepackage{xcolor}         
\usepackage{enumitem}
\setlist[itemize]{leftmargin=0.5cm}    
\setlist[enumerate]{leftmargin=0.5cm}
\usepackage{graphicx}
\usepackage{makecell}  
\usepackage{adjustbox} 
\usepackage{amsmath}
\usepackage{amssymb}
\usepackage{mathtools}
\usepackage{amsthm}
\usepackage{multirow,hhline}
\usepackage{caption}
\usepackage{wrapfig}
\usepackage[capitalize,noabbrev]{cleveref}
\usepackage{mdframed}
\usepackage{listings}
\usepackage{subfigure}
\usepackage{colortbl}
\usepackage{arydshln}
\usepackage{fancyvrb}
\usepackage{tabularx}  %
\usepackage{placeins} 
\newcommand{\mymodel}{\textit{Table-r1}}

\title{\mymodel~: Self-supervised and Reinforcement Learning for Program-based Table Reasoning in Small Language Models}

\author{
  Rihui Jin\textsuperscript{1,}\textsuperscript{2},
  Zheyu Xin\textsuperscript{1},
  Xing Xie\textsuperscript{1},
  Zuoyi Li\textsuperscript{1},
  Guilin Qi\textsuperscript{1}\thanks{Corresponding Author}, \\
  \textbf{Yongrui Chen}\textsuperscript{1},
  \textbf{Xinbang Dai}\textsuperscript{1},
  \textbf{Tongtong Wu}\textsuperscript{2},
  \textbf{Gholamreza Haffari}\textsuperscript{2} \\
  \textsuperscript{1}Southeast University, Nanjing, China \\
  \textsuperscript{2}Monash University, Australia \\
  \texttt{\{ari\_king, gqi, yrchen, xbd\}@seu.edu.cn} \\
  \texttt{\{xinzheyu464, xingxie.cn, lzu125808\}@gmail.com} \\
  \texttt{\{Tongtong.wu, Gholamreza.Haffari\}@monash.edu}
}

\begin{document}

\maketitle

\begin{abstract}
Table reasoning (TR) requires structured reasoning over semi-structured tabular data and remains challenging, particularly for small language models (SLMs, e.g., LLaMA-8B) due to their limited capacity compared to large LMs (LLMs, e.g., GPT-4o). 
To narrow this gap, we explore program-based TR (P-TR), which circumvents key limitations of text-based TR (T-TR)—notably in numerical reasoning—by generating executable programs. 
However, applying P-TR to SLMs introduces two challenges: (i) vulnerability to heterogeneity in table layouts, and (ii) inconsistency in reasoning due to limited code generation capability.
We propose \textit{Table-r1}, a two-stage P-TR method designed for SLMs. 
Stage 1 introduces an innovative self-supervised learning task, Layout Transformation Inference, to improve tabular layout generalization from a programmatic view.
Stage 2 adopts a mix-paradigm variant of Group Relative Policy Optimization, enhancing P-TR consistency while allowing dynamic fallback to T-TR when needed.
Experiments on four TR benchmarks demonstrate that \textit{Table-r1} outperforms all SLM-based methods, achieving at least a 15\% accuracy improvement over the base model (LLaMA-8B) across all datasets and reaching performance competitive with LLMs. 
\footnote{Source code will be shown in \url{https://github.com/AriKing11/Table_r1_public}} 
\vspace{-0.3cm}
\end{abstract}

\definecolor{mygray}{gray}{.91}

\section{Introduction}
\label{section: intro}

Table reasoning (TR)~\citep{survey2024tablereasoning, hegta} is a fundamental yet challenging task, requiring structured reasoning over the two-dimensional semantics of tabular data~\citep{tabprompt}. 
With the emergence of large language models (LLMs), TR has seen notable progress~\citep{survey2024llmtable}.
Existing LLM-based methods fall into two main paradigms: text-based TR (T-TR), e.g., Table-CoT~\citep{Table-CoT}, which treats tables as plain text and generates answers directly; and program-based TR (P-TR), e.g., Binder~\citep{Binding}, which generates executable programs to derive answers. Subsequent studies~\citep{tama, Tree-of-Table, graphotter} have further advanced both approaches.

\begin{figure}[t]
  \centering
  \includegraphics[width=0.75\textwidth]{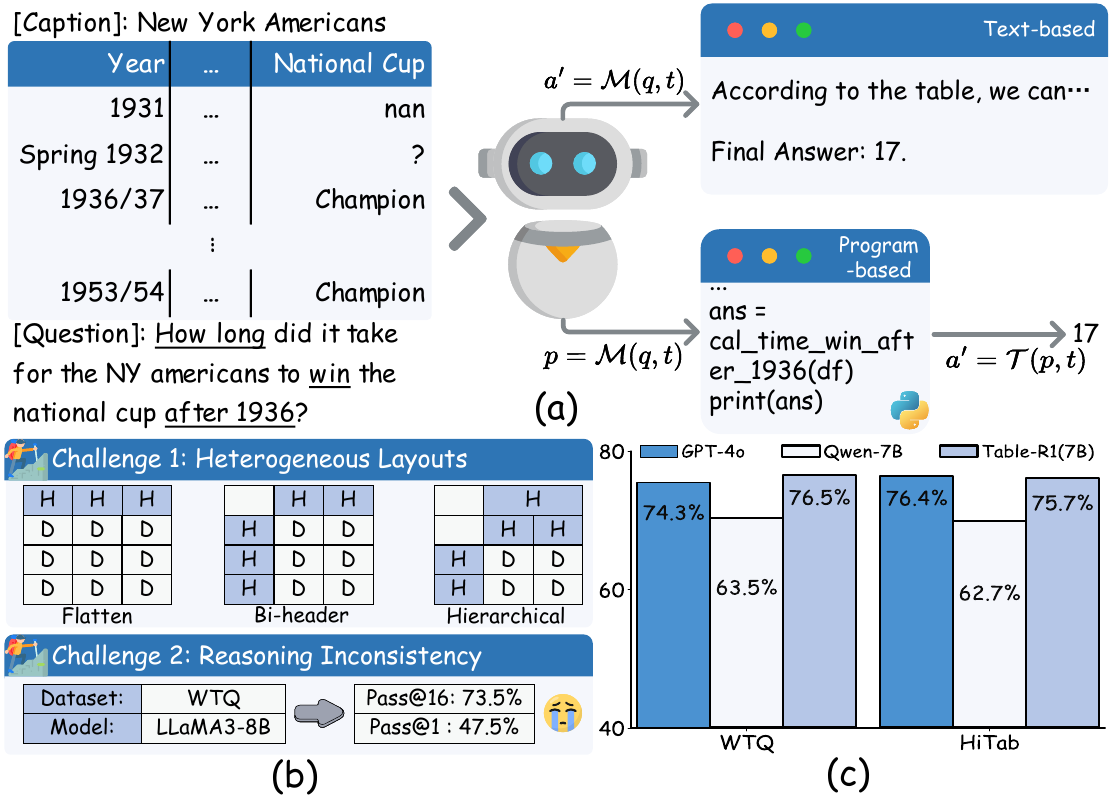}
  \caption{(a) Two TR paradigms: the text-based directly generates answers, while the program-based produces executable code (e.g., Python) to derive answers via a code executor.
(b) Two challenges for SLMs in P-TR: heterogeneous layouts (e.g., complex and flexible header structures) hinder accurate indexing, and limited reasoning ability causes inconsistent performance.
(c) Performance comparison on two TR benchmarks: LLMs(GPT-4o) outperform SLMs (Qwen-7B) by a large margin; \mymodel{}, built on Qwen-7B, significantly narrows the gap and rivals LLM performance.}
  \label{fig: intro}
  \vspace{-0.3cm}
\end{figure}

However, when moving from LLMs (e.g., GPT-4o) to small language models (SLMs, e.g., Qwen-7B)~\cite{srivastava2025towards}, performance drops significantly (Fig.~\ref{fig: intro}(c)), primarily due to SLMs’ weaker reasoning and text comprehension abilities.
While prior efforts~\citep{tablellama, tart, tablegpt2} have fine-tuned SLMs under the T-TR setting, these face inherent limits such as restricted context windows and poor numerical handling~\citep{chen2022program}. 
In contrast, P-TR methods naturally mitigate these issues~\citep{mixSC, tart}\footnote{Further discussed in~\S\ref{section: error_study}}. 
This motivates a central question: \textbf{Can we design a P-TR method enabling SLMs to match or even surpass LLM performance?}

Pursuing this direction, however, requires overcoming two key challenges of SLMs under the P-TR paradigm, as highlighted in Fig.~\ref{fig: intro}(b):
(i) poor generalization to heterogeneous layouts (flat, bi-level, hierarchical headers), which impairs accurate referencing of contents via headers as arguments in programs; and
(ii) reasoning inconsistency, as evidenced by high pass@16 (73.5\%) but low pass@1 (47.5\%) in single-shot settings.
To address them, we propose \mymodel{}, a two-stage P-TR method designed for SLMs:
In Stage 1, we introduce an innovative self-supervised learning task—Layout Transformation Inference—to enhance layout generalization by training the model to infer structural transformations from a programmatic perspective. This promotes accurate header usage and content referencing.
In Stage 2, we adopt and extend Group Relative Policy Optimization (GRPO)~\citep{deepseekmath} into a mix-paradigm variant that prioritizes P-TR while selectively leveraging T-TR as a fallback. The model dynamically switches strategies based on the context and prior reasoning trace, guided by several TR-specific rewards. GRPO then optimizes the model's policy toward higher-reward completions, enhancing both robustness and adaptability.
Extensive experiments across four public TR benchmarks demonstrate that \mymodel{} consistently surpasses all prior SLM-based methods and achieves performance on par with LLMs.

The contributions of this paper can be summarized as follows:
\begin{enumerate}
    \item We introduce an innovative self-supervised learning task tailored for the P-TR, designed to improve SLMs’ ability to understand heterogeneous table layouts without manual annotation.
    \item To the best of our knowledge, we are the first to apply GRPO to TR. We further adapt it into a mix-paradigm GRPO, design task-specific reward functions, and conduct an in-depth analysis of GRPO's effectiveness in the TR.
    \item Extensive experiments demonstrate that \mymodel{} consistently outperforms all existing SLM-based methods and achieves performance competitive with that of LLMs across multiple TR benchmarks.
\end{enumerate}

\section{Related Work}
\subsection{LM-based Table Reasoning}

Recent advances in LM-based TR follow two main paradigms—Text-based TR and Program-based TR—as illustrated in Fig.~\ref{fig: intro}(a) and detailed in \S\ref{section: problem_statement}.

\textbf{Text-based Paradigm.}  
In this paradigm, the LLM takes the table and the corresponding question as input and directly generates the final answer. 
\citet{Table-CoT} first demonstrated the potential of LLMs in TR by leveraging COT prompting~\citep{cot} and ICL~\citep{gpt3}. 
Building on this, subsequent works~\citep{bowenZhao, Table-LLM-Specialist, tableformatsurvey} introduced techniques such as parse-tree decoding, the ReAct~\citep{react}, and format-aware table transformations to improve reasoning performance. 
However, such prompt-based methods struggle to generalize to SLMs, whose capabilities are significantly weaker than those of LLMs.
Although several studies~\citep{tablellama, tama, tablegpt, tablegpt2, Structlm, Tablelora} have explored instruction tuning to bridge this gap, the performance of SLMs remains limited, primarily due to context length constraints. Truncated tables often omit critical information required for accurate reasoning. 
Motivated by these limitations, our work takes an alternative direction to unlock the potential of SLMs better, enabling them to excel in TR tasks.

\textbf{Program-based Paradigm.}  
This paradigm prompts the LLM to generate executable code, typically in Python, based on the provided table and question, and subsequently executes the code to derive the final answer.
Binder~\citep{Binding} was the first to introduce the Program-of-Thought framework~\citep{chen2022program, gao2023pal} to TR tasks. These methods leverage an external code executor to overcome limitations in context length and numerical reasoning commonly observed in LLMs~\citep{schick2024toolformer, llmmath}.
As a result, P-TR methods have gained substantial traction, with follow-up works~\citep{ReAcTable, chain-of-table, dater, Tree-of-Table, alter, trove} integrating techniques such as ReAct, Tree-of-Thoughts~\citep{TreeOfThoughts}, and graph-based modeling to enhance reasoning capabilities further. 
Notably, some methods~\citep{mixSC, yangtriples} have adopted P-TR as the primary framework, with text-based paradigms playing a supplementary role to improve robustness and coverage.
However, most of these methods remain heavily dependent on the code generation abilities of LLMs, posing significant challenges when adapting to SLMs. 
Although some initial efforts~\citep{trove, tart} have explored P-TR in the context of small models, a considerable performance gap still exists. 
In this work, we build upon the P-TR paradigm and propose an innovative method that enables SLMs to achieve performance comparable to that of their larger counterparts.
\vspace{-0.2cm}
\subsection{Reinforcement Learning for LLM-based Reasoning}
\vspace{-0.2cm}
Recently, a growing body of research has focused on enhancing the reasoning capabilities of LLMs. 
The impressive performance of o1~\citep{ali-o1} in complex reasoning tasks has spurred the academic community to focus on optimizing models through RL. 
Notably, DeepSeek-R1~\citep{r1} and Kimi-1.5~\citep{kimi1.5} have both utilized advanced versions of Proximal Policy Optimization~\citep{ppo}, resulting in significant improvements in LLM performance on verifiable tasks, such as mathematics and code generation. 
Among these, the GRPO~\citep{deepseekmath} algorithm employed by DeepSeek-R1 has been widely validated for its effectiveness in boosting the reasoning abilities of SLMs~\citep{Understanding-r1-zero-like, zeng2025simplerl, yue2025does}. Despite these advancements, to date, no work has explicitly applied RL-based methods to the domain of TR.

\section{Preliminaries}

\subsection{Problem Statement}
\label{section: problem_statement}
In the TR task, each data sample is formalized as a triplet $\langle t, q, a \rangle$, where $t \in \mathbb{T}$ denotes a semi-structured table, $q \in \mathbb{Q}$ represents a natural language question, and $a \in \mathbb{A}$ is the ground-truth answer derived from table $t$.
A semi-structured table $t$ is composed of a header $H$ and a body section $D$. 
The header $H$ encodes schema-level semantics and is designed for readability.
Unlike flat database schemas, it may have a hierarchical layout or be split into top and left headers, especially in financial reports~\citep{hitab, aitqa}.
The goal of models in TR is to learn a mapping function $ \phi: \mathbb{Q} \times \mathbb{T} \rightarrow \mathbb{A} $, which outputs a correct answer $a'$ based on the input pair $(q, t)$.
In T-TR, the reasoning process is directly handled by a language model $ \mathcal{M} $, which predicts the answer in a single step:
\begin{equation}
    a' = \mathcal{M}(q, t),
\end{equation}
where $a'$ is the model's generated response.
Alternatively, P-TR paradigm decomposes the TR into two stages: the model $ \mathcal{M} $ first produces an intermediate program $p$, which is then executed by an external executor $ \mathcal{E} $ to obtain the final answer:
\begin{align}
    p = \mathcal{M}(q, t), \quad 
    a' = \mathcal{E}(p, t).
\end{align}
In our method, the program $p$ corresponds to Python code.

\section{Methodology}
\label{section: method}
\begin{figure}[!t]
    \centering
    \includegraphics[width =1.0\linewidth]{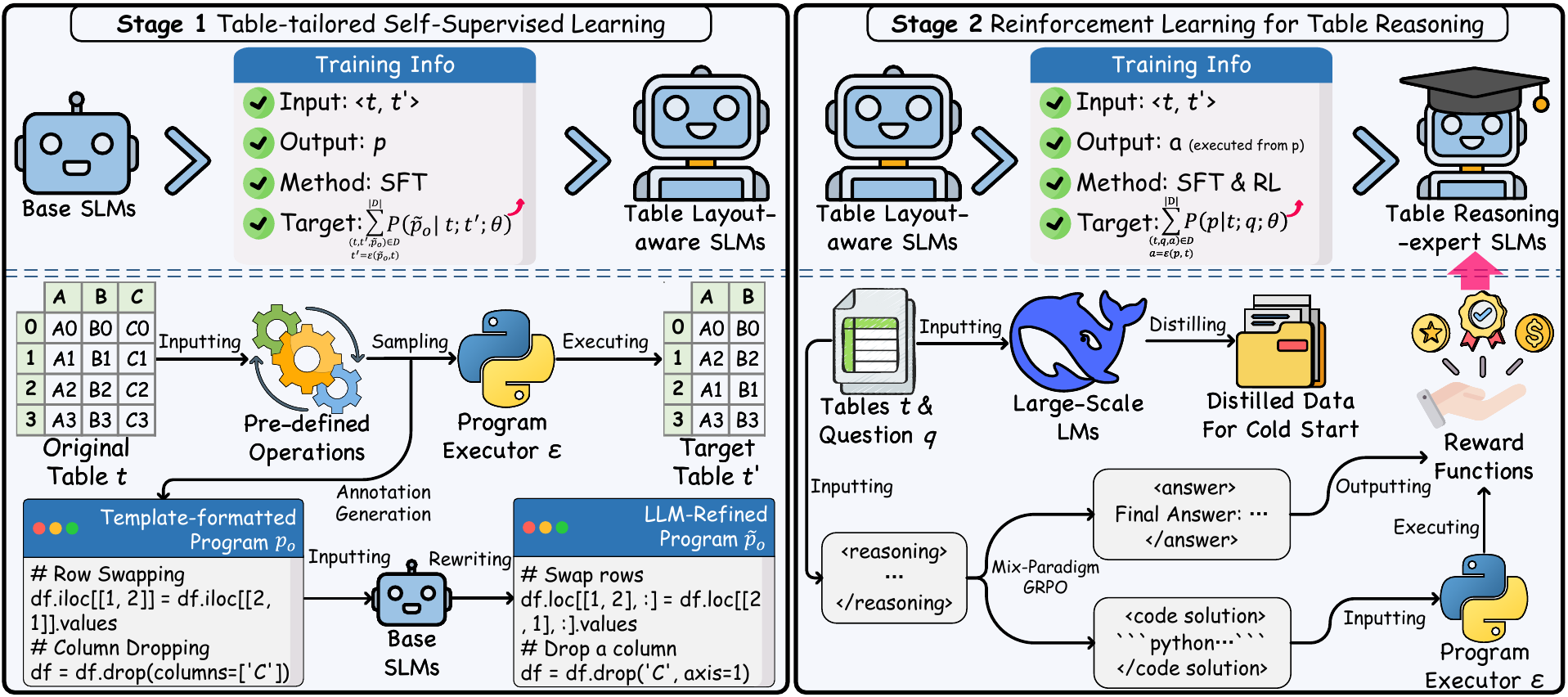}
    \caption{Overview of \mymodel{}. Stage~1 introduces an SSL task to improve layout understanding, with annotations auto-generated (bottom-left). Stage 2 applies RL, starting from teacher-guided cold start and followed by a mix-paradigm GRPO, enabling dynamic strategy selection. Reward functions provide numerical feedback based on outputs.}
    \label{fig: overview}
    \vspace{-0.4cm}
\end{figure}    

We propose \mymodel{}, a two-stage training method for enhancing TR. As illustrated in Fig.~\ref{fig: overview}, Stage~1 (\S\ref{subsection: stage1}) strengthens the model’s understanding of tabular layouts and semantics; Stage~2 (\S\ref{subsection: stage2}) enhances code-based reasoning via reinforcement learning.

\subsection{Stage 1: Self-Supervised Learning Tailored for Program-based Table Reasoning}
\label{subsection: stage1}
Real-world tables often deviate from standard formats, exhibiting diverse layouts—e.g., flat, hierarchical, bi-directional headers—that hinder semantic parsing and accurate indexing.

To mitigate this issue, we propose a self-supervised task (SSL) task called Layout Transformation Inference (LTI), which trains the model to compare two tables—an original table $t$ and its transformed counterpart $t'$—and generate a program $\tilde{p}_o$ that reconstructs $t'$ from $t$:
\begin{equation}
    t' = \mathcal{E}(\tilde{p}_o, t),
\end{equation}
where $\mathcal{E}$ is a program executor.  
LTI demands fine-grained comparison between $t$ and $t'$ to identify layout changes and express them via executable transformations. This encourages the model to distinguish headers from data and use headers as structural anchors. The task is non-trivial—GPT-4o achieves only 45\% accuracy in ICL, highlighting the difficulty and potential for improving the model's understanding ability of layouts without manual annotation.

\subsubsection{Training Data Synthesis}
\textbf{Synthetic Instance Construction}  
To automatically generate target tables, we define a set of transformation operations $\mathcal{O} = \{o_1, o_2, \dots, o_K\}$, where each $o_i$ is a deterministic function modifying the layout of a table $t$ (e.g., row swapping, column deletion; see Appx.~\ref{appx: operations} for details).  
For each $t$, we apply a random sequence of $n$ operations ($1 \leq n \leq 3$) to generate a transformed table $t'$:
\begin{align}
    p_o = o_n \circ \cdots \circ o_1, \quad t' = \mathcal{E}(p_o, t),
\end{align}
where $p_o$ represents the composition of applied operations and serves as the supervision labels.

\textbf{LLM-Based Label Rewriting}  
Directly using template-based programs $p_o$ in SSL can introduce distributional mismatch, causing catastrophic forgetting~\citep{wu2024continual}.  
To mitigate this, we adopt a continual learning strategy based on self-distillation~\citep{self-distilled}.  
Specifically, the base model $\mathcal{M}_{\text{base}}$ rewrites the raw synthetic program $p_o$ into a semantically equivalent but more fluent version $\tilde{p}_o$:
\begin{equation}
    \tilde{p}_o = \mathcal{M}_{\text{base}}(t, t', p_o),
\end{equation}which serves as the final training label.

\subsubsection{Fine-Tuning Procedure}  
We fine-tune the model on synthesized triplets $(t, t', \tilde{p}o)$ using the following objective:
\begin{equation}
\mathcal{L}_{\text{LTI}} = \frac{1}{|D|} \sum_{\substack{(t, t', \tilde{p}_o) \in D \\ \mathcal{E}(\tilde{p}_o, t) = t'}} 
\left[ -\log P(\tilde{p}_o \mid t; t'; \theta) \right],
\end{equation}
where $D$ is the training set and $\theta$ the model parameters. The loss is computed only when $\tilde{p}_o$ faithfully transforms $t$ into $t'$.

\subsection{Stage 2: Reinforcement Learning for Table Reasoning}
\label{subsection: stage2}
We enhance code-based reasoning via RL. GRPO~\citep{r1} is particularly well-suited for SLMs due to its efficiency and strong performance on verifiable tasks, outperforming traditional methods such as PPO~\citep{ppo}. 
Meanwhile, prior work~\citep{mixSC, yangtriples} shows that T-TR can complement P-TR under specific contexts. 
Motivated by these insights, we integrate a mix-paradigm variant of GRPO into \mymodel{}, enhancing P-TR consistency while allowing dynamic fallback to T-TR when needed.
We now detail the components: completion templates (\S\ref{section: templates}), distillation-based cold start (\S\ref{section: cold-start}), reward functions (\S\ref{section: reward}), and the mix-paradigm GRPO~(\S\ref{section: grpo}).

\subsubsection{Completion Templates}
\label{section: templates}
As shown in Fig.~\ref{fig: overview}, we adopt two structured output formats. 
Both prompt the model to generate intermediate reasoning before deciding on an answer format. 
If the P-TR template is selected, the answer is enclosed in \texttt{<code\_solution>} tags containing Python code; otherwise, a direct answer appears within \texttt{<answer>} tags. 
This design, inspired by~\citep{Scratchpads}, encourages deliberate reasoning and supports flexible strategy selection. 
Examples are provided in Appx.~\ref{appx: prompt_grpo}.

\subsubsection{Cold Start with Distilled Data}
\label{section: cold-start}
Inspired by~\cite{Understanding-r1-zero-like}, we adopt a distillation-based cold start strategy before RL. 
We employ LLMs~\citep{deepseekai2025deepseekv3technicalreport} as teacher models to generate reasoning traces and code solutions on the training split, following the prompt format in Appx.~\ref{appx: prompt_p}. 
The distilled data is then used to perform SFT on the base model.

\subsubsection{Reward Functions}
\label{section: reward}
Reward design is central to RL~\citep{Understanding-r1-zero-like}. In GRPO, reward functions evaluate each model-generated completion, guiding optimization toward better outputs. Formally, we define:
\begin{equation}
R(c_i) = \sum_j R_j(c_i, t, a), \quad
\end{equation}
where $R_j$ denotes the $j$-th reward component, and $R(c_i)$ is the total reward for completion $c_i$.

Beyond format-adherence reward functions (see Table~\ref{table: reward_functions} for more details), the following reward functions $R_j$ are introduced  to encourage more reliable and interpretable generation:
\begin{itemize}

\item \textbf{Compilation Correctness.} 
Program-based outputs receive the reward only when the code compiles successfully. Text-based outputs are assumed valid by default, which introduces a bias that is counteracted by other reward terms.

\item \textbf{Answer Correctness.}
For P-TR outputs, we execute the generated code to obtain the answers; for text-based outputs, we use regular expression matching. To refine reward granularity, we compare model outputs with golden answers using similarity metrics. 
Incorrect but compilable outputs (e.g., \texttt{print(final answer)}) are penalized heavily.
We also penalize the use of text-based answers on questions requiring P-TR (e.g., extremum queries over truncated tables).

\item \textbf{Short Code Penalty.} 
To discourage degenerate behaviors (e.g., overly short programs), we impose penalties on code completions that are below a specified length threshold.

\end{itemize}

\subsubsection{Mix-paradigm GRPO}
\label{section: grpo}
We extend GRPO to a \textit{mix-paradigm} variant, allowing the model to prioritize P-TR while flexibly falling back to T-TR when appropriate. GRPO optimizes a policy by maximizing the relative advantage of high-reward completions, regularized against a reference policy.
Given a query-table pair $\langle q, t \rangle$, the model generates multiple completions $c_i = [s; z']$, where $s$ is a reasoning trace and $z'$ is the final answer in a specific format.
We define the output space $\mathcal{Z} = {\text{program}, \text{text}}$, where each $z \in \mathcal{Z}$ denotes a distinct answer form—executable code or direct output.
The model implicitly selects $z'$ during generation, conditioned on $s$:
\begin{equation}
z' = \arg\max_{z \in \mathcal{Z}} P(z \mid s; \theta).
\end{equation}

Each $c_i$ is then evaluated using a TR-specific reward function $R_{c_i}$ (defined in \S\ref{section: reward}). 
We compute its relative advantage within the batch as:
\begin{equation}
A(c_i) = \frac{R(c_i) - \mu_R}{\sigma_R + \epsilon},
\end{equation}
where $\mu_R$ and $\sigma_R$ are the mean and standard deviation of rewards among the sampled completions ${c_i}$ in the same batch, and $\epsilon$ is a small constant for numerical stability.
The policy is updated by maximizing a clipped surrogate objective:
\begin{equation}
\mathcal{L}_{\text{GRPO}} =
\mathbb{E}_{c_i \sim \pi_\theta} \left[
\min\left( r(c_i) A(c_i), \ \text{clip}(r(c_i), 1 - \epsilon, 1 + \epsilon) A(c_i) \right)
\right] - \lambda \cdot \text{KL}(\pi_\theta || \pi_{\theta_{\text{ref}}}),
\end{equation}
where $r(c_i) = \pi_\theta(c_i) / \pi_{\theta_{\text{ref}}}(c_i)$ is the likelihood ratio, and $\lambda$ controls the KL regularization strength. This encourages exploration of higher-reward answers while constraining divergence from the reference policy.

\section{Experiments}
In this section, we first present the datasets and implementation details. We then report experimental results comparing our method against strong baselines, including SOTA, to validate its effectiveness. Additionally, we explore several key findings that arise from our study, which we formulate into the following research questions:
\begin{itemize}[leftmargin=1cm]

\item \textbf{RQ1}: For SLMs, what are the respective strengths and weaknesses of the T-TR and P-TR paradigms in handling real-world TR tasks?
\item \textbf{RQ2}: How does each module contribute to the overall performance gains in TR, and in which scenarios does mix-paradigm GRPO demonstrate clear advantages?
\item \textbf{RQ3}: What are the respective roles and impacts of SFT and RL in improving the reasoning capability of P-TR models?
\item \textbf{RQ4}: To what extent does organizing training data by difficulty levels influence the learning efficiency and final performance of GRPO?

\end{itemize}


\subsection{Experimental Settings}
\label{section: settings}
\begin{wraptable}{r}{0.5\textwidth}
  \centering
  \vspace{-1em}
  \resizebox{0.5\textwidth}{!}{
    \begin{tabular}{lcccc}
      \toprule
      \textbf{Dataset} & \textbf{Header} & \textbf{\# $t$} & \textbf{\# $q$} & \textbf{Domain} \\
      \midrule
      WTQ~\citep{wtq}         & Flat  & 2.1k   & 22k   & Geneal \\
      TabFact~\citep{TabFact} & Flat  & 16k    & 118k  & General \\
      HiTab~\citep{hitab}     & Hierarchical    & 3.5k   & 10k   & Crime, Health \\
      AIT-QA~\citep{aitqa}    & Hierarchical    & 116    & 515   & Airline \\
      \bottomrule
    \end{tabular}
  }
  \caption{Dataset statistics summary}
  \label{table: Dataset_Statistics}
  \vspace{-1em} 
\end{wraptable}

\textbf{Datasets \& Metrics \& Implementation.}
We evaluate on four widely used TR benchmarks. Dataset statistics are summarized in Table~\ref{table: Dataset_Statistics}. For all datasets except AIT-QA, we use the original train-test splits; AIT-QA is randomly split 8:2 based on questions.  Following prior work~\citep{mixSC, Binding}, we adopt self-consistency (SC) with majority voting, using 5 samples per prediction, and report exact match accuracy as the evaluation metric.
Implementation details can be found in Appx.~\ref{appx: Implementation}.

\textbf{Baselines.}
We compare against recent TR methods (see Table~\ref{table: main-results}), covering both text-based and program-based paradigms. Additionally, we evaluate the text-based and program-based performance of SOTA LLMs (GPT-4o~\citep{jaech2024openai}, DeepSeek-v3~\citep{deepseekai2025deepseekv3technicalreport}, DeepSeek-R1~\citep{r1}) and base models (Qwen2.5-Coder-7B~\citep{qwen2_5}, LLaMA3.1-8B-Instruct~\citep{llama3}) under the same prompt settings with SC=5 as in \mymodel{}.

\newcommand{\txt}{\cellcolor[rgb]{0.89,0.93,1.0}\texttt{T-TR}}
\newcommand{\symb}{\cellcolor[rgb]{0.92,0.95,0.91}\texttt{P-TR}}
\begin{table*}[ht]
\centering
\small
\renewcommand{\arraystretch}{1.2}
\setlength{\tabcolsep}{5pt}

\begin{adjustbox}{width=0.95\textwidth}

\begin{tabular}{cllcccccc}
\toprule
\multicolumn{2}{c}{\multirow{2}{*}{\textbf{Method}}} 
& \multirow{2}{*}{\textbf{Base Model}} 
& \multirow{2}{*}{\textbf{Paradigm}} 
& \multicolumn{2}{c}{\textbf{Flatten Tables}} 
& \multicolumn{2}{c}{\textbf{Hierarchical Tables}}  \\

\cmidrule(lr){5-6} \cmidrule(lr){7-8} 
& & & & WTQ & TabFact & HiTab & AIT-QA  \\ 
\midrule

\rowcolor[rgb]{0.91,0.92,0.89}
\multicolumn{9}{c}{\textit{Large-Scale LLMs}} \\

& E2E (one-shot)                        & GPT-4o                & \txt   & 63.2          & 87.6             & 82.4              & 88.5   \\ 
& E2E (one-shot)                        & DeepSeek-v3 (670B)    & \txt   & 61.0          & 84.5             & \textbf{85.3}     & \textbf{91.7} \\
& E2E (one-shot)                        & GPT-4o                & \symb  & 74.3          & 88.1             & 76.4              & 86.8       \\ 
& E2E (one-shot)                        & DeepSeek-v3 (670B)    & \symb  & 75.0          & 87.2             & 77.0              & 87.0       \\ 
& E2E (one-shot)                        & DeepSeek-r1 (670B)    & \symb  & \textbf{82.0} & \textbf{93.0}    & 82.3              & 90.0        \\ 

\hdashline
& TableParser~\citep{zhao2023large}     & Qwen2-72B-Inst        & \txt   & --    & --    & 44.6  & 64.9     \\
& Table-Critic~\citep{Table-critic}     & Qwen2-72B-Inst        & \txt   & 77.2  & 92.6  & --    & --       \\
& GraphOTTER~\citep{graphotter}         & Qwen2-72B-Inst        & \symb  & --    & --    & 73.7  & 88.3     \\
& PoTable~\citep{mao2024potable}        & LLaMA3.1-70B-Inst     & \symb  & 65.6  & 87.1  & --    & --       \\
& TableMaster~\citep{tablemaster}       & LLaMA3.1-70B-Inst     & \symb  & 78.0  & 91.2  & --    & --       \\
& E5~\citep{e5}                         & GPT-4                 & \symb  & 65.5  & 88.8  & 85.1  & --       \\
& Norm-DP\&Agent~\citep{mixSC}          & GPT-3.5               & \symb  & 73.7  & 88.5  & --    & --       \\
& TIDE-DP\&Agent~\citep{yangtriples}    & GPT-3.5               & \symb  & 75.0  & 89.8  & --    & --       \\

\midrule
\midrule

\rowcolor[rgb]{0.91,0.92,0.89}
\multicolumn{9}{c}{\textit{Small-Scale LLMs}} \\

& E2E  (one-shot)                       & LLaMA3.1-8B-Inst           & \symb  & 55.3  & 65.2  & 49.2      & 48.8        \\ 
& E2E  (one-shot)                       & Qwen2.5-Coder-7B-Inst      & \symb  & 63.5  & 76.0  & 62.7      & 55.5       \\ 

\hdashline

& TableLLama~\citep{tablellama}         & LLaMA2-7B             & \txt   & --    & 82.6  & 60.5  & --        \\ 
& TAMA~\citep{tama}                     & LLaMA3.1-8B-Inst      & \txt   & 52.9  & 73.8  & 63.5  & 89.2      \\ 
& TableGPT2~\citep{tablegpt2}           & Qwen2.5-7B            & \txt   & 61.42 & 77.8  & 70.3  & --        \\
& TableLoRA~\citep{Tablelora}           & LLaMA3-8B-Inst        & \txt   & 53.5  & 84.0  & 58.6  & --        \\
& Structlm~\citep{Structlm}             & Mistral-7B            & \txt   & 56.8  & 84.6  & --    & --        \\
& Tart~\citep{tart}                     & DeepSeek-7B           & \symb  & --    & 71.3  & --    & --        \\

\midrule

& \mymodel{} & Qwen2.5-Coder-7B-Inst & \symb  
& $\textbf{76.5} \pm 0.74$  
& $\textbf{85.3} \pm 0.81$  
& $\textbf{76.7} \pm 0.72$  
& $\textbf{89.9} \pm 0.88$ \\
& \mymodel{} & LLaMA3.1-8B-Inst & \symb  
& $74.3 \pm 0.77$  
& $84.4 \pm 0.85$  
& $74.2 \pm 1.03$  
& $86.5 \pm 0.93$ \\

\bottomrule
\end{tabular}
\end{adjustbox}

\caption{Main results of LLM-based methods across various TR datasets. \textbf{Paradigm} indicates the modeling style: \txt{} (text-based) and \symb{} (program-based). The evaluation metric is accuracy. The E2E (one-shot) series share the same prompts as Table-R1, with one-shot demonstration included. Averaged over three runs; $\pm$ denotes standard deviation.}
\vspace{-0.7cm}
\label{table: main-results}
\end{table*}

\subsection{Overall Performance}
Table~\ref{table: main-results} reports results of Table-R1 against all baselines across four TR benchmarks. Our method consistently outperforms all SLMs of comparable scale and matches the performance of top-tier LLMs.
Several key observations emerge:
(i) For LLMs, P-TR performs better on flattened tables (e.g., WTQ, TabFact), whereas T-TR is more effective on hierarchical tables (e.g., HiTab, AIT-QA); see \S~\ref{section: error_study} for analysis.
(ii) Models like Table-R1 and v3, which operate without external plugins, achieve the best results—highlighting the critical role of the base model’s capacity in TR.
That said, proper strategies still provide substantial gains, even for smaller models.

\subsection{Error Studies (RQ1)}
\label{section: error_study}
\begin{wrapfigure}{r}{0.5\textwidth}
  \vspace{-0.15cm}
  \centering
  \includegraphics[width=0.5\textwidth]{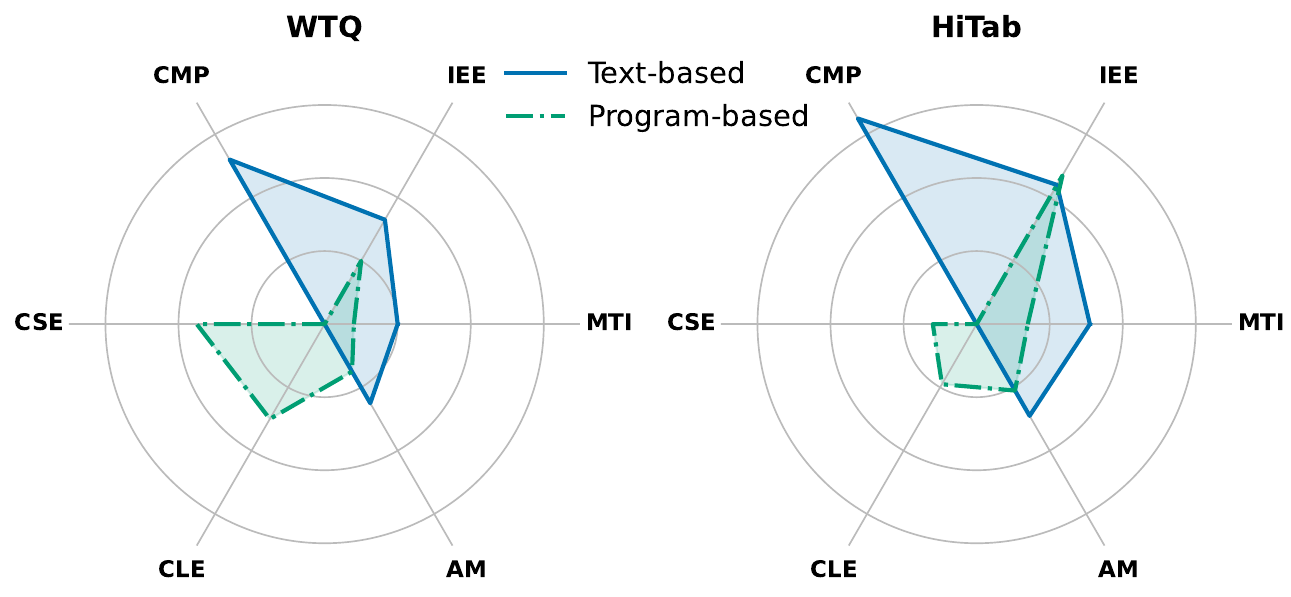}
  \caption{Error type distribution.}
  \label{fig: radar}
\end{wrapfigure}

To analyze the limitations of the two paradigms (RQ1), we manually inspect 200 instance outputs each from WTQ and HiTab using Qwen2.5-7B-Instruct (text-based) and Qwen2.5-Coder-7B-Instruct (program-based). The program-based model is restricted to the first 10 rows due to token limits, while the text-based model processes the entire table (truncated if exceeding 4096 tokens). 
We follow the classification scheme in~\citep{mixSC, e5}, assigning each prediction to one of six error categories: 

\quad • Missing Table Information (MTI):  The answer-relevant content is truncated and unavailable to the model.

\quad • Information Extraction Errors (IEE): Caused by structural heterogeneity, manifesting as hallucinations or index misuse (Examples shown in Fig.~\ref{fig: case1} and Fig.~\ref{fig: case2}).

\quad • Computation Errors (CMP): Errors stemming from incorrect numerical reasoning.

\quad • Code Syntax Errors (CSE): The generated code fails to compile due to syntactic issues.

\quad • Logical Errors in Code (CLE): The code compiles and runs but yields incorrect results due to flawed reasoning or filtering logic. (An example shown in Fig.~\ref{fig: case3}).

\quad • Answer Misalignment (AM): where the output format diverges from the expected answer.
CSE and CLE differ from IEE in that they assume syntactic or semantic code validity, but not necessarily accurate indexing.

As shown in Fig.~\ref{fig: radar}, computation errors are most frequent in text-based methods, reflecting LLMs' known weaknesses in arithmetic~\citep{gao2023pal}. These models also struggle with large tables, where high information density increases both extraction errors and the risk of answer truncation due to context limits.
The program-based paradigm alleviates these issues but still suffers from syntactic and logical flaws in code generation. In particular, HiTab's complex hierarchical indexing significantly raises the information extraction errors that occur in the program.

\subsection{Ablation Study (RQ2, RQ3)}

\begin{table}[t]
\centering
\small
\begin{tabular}{@{}clllll@{}}
\toprule
\multicolumn{1}{c}{} & \textbf{Module} & \textbf{WikiTQ} & \textbf{TabFact} & \textbf{HiTab} & \textbf{AiT-QA} \\
\midrule
\multirow{6}{*}{\rotatebox{90}{\textbf{Qwen-7B}}} 
  & \textit{\mymodel{} (w/ all modules)}   & \textbf{76.5} & \textbf{85.3} & \textbf{76.7} & \textbf{89.9} \\
  & \textit{\mymodel{} (w/o all modules)}  & 63.5 ($\downarrow$13.0) & 76.0 ($\downarrow$9.3) & 62.7 ($\downarrow$14.0) & 55.5 ($\downarrow$34.4) \\
  & \;\; - SSL                             & 74.2 ($\downarrow$2.3)  & 84.1 ($\downarrow$1.2) & 74.4 ($\downarrow$2.3)  & 87.3 ($\downarrow$2.6) \\
  & \;\; - Cold Start                      & 71.1 ($\downarrow$5.4)  & 80.5 ($\downarrow$4.8) & 68.8 ($\downarrow$7.9)  & 67.3 ($\downarrow$22.6) \\
  & \;\; - GRPO                            & 71.7 ($\downarrow$4.8)  & 81.2 ($\downarrow$4.1) & 70.4 ($\downarrow$6.3)  & 83.3 ($\downarrow$6.6) \\
  & \;\; - Mix                             & 75.1 ($\downarrow$1.4)  & 84.4 ($\downarrow$0.9) & 73.6 ($\downarrow$3.1)  & 86.4 ($\downarrow$3.5) \\
\midrule
\multirow{6}{*}{\rotatebox{90}{\textbf{LLaMA-8B}}} 
  & \textit{\mymodel{} (w/ all modules)}   & \textbf{74.3} & \textbf{84.4} & \textbf{74.2} & \textbf{86.5} \\
  & \textit{\mymodel{} (w/o all modules)}  & 55.3 ($\downarrow$19.0) & 65.2 ($\downarrow$19.2) & 49.2 ($\downarrow$25.0) & 48.8 ($\downarrow$37.7) \\
  & \;\; - SSL                             & 72.2 ($\downarrow$2.1)  & 83.0 ($\downarrow$1.4) & 70.9 ($\downarrow$3.3)  & 83.4 ($\downarrow$3.1) \\
  & \;\; - Cold Start                      & 67.5 ($\downarrow$6.8)  & 77.3 ($\downarrow$7.1) & 63.8 ($\downarrow$10.4) & 63.5 ($\downarrow$23.0) \\
  & \;\; - GRPO                            & 70.0 ($\downarrow$4.3)  & 77.4 ($\downarrow$7.0) & 66.1 ($\downarrow$8.1)  & 77.3 ($\downarrow$9.2) \\
  & \;\; - Mix                             & 73.2 ($\downarrow$1.1)  & 83.2 ($\downarrow$1.2) & 69.6 ($\downarrow$4.6)  & 81.4 ($\downarrow$5.1) \\
\bottomrule
\end{tabular}
\vspace{0.1cm}
\caption{Ablation results on four TR benchmarks with Qwen-7B and LLaMA-8B. Removing each module consistently degrades performance, verifying their complementary contributions.}
\vspace{-0.5cm}
\label{table: ablation}
\end{table}

\begin{wraptable}{r}{0.5\textwidth}
  \centering
  \vspace{-1em}
  \resizebox{0.48\textwidth}{!}{
    \begin{tabular}{lcccc}
      \toprule
        \textbf{Module} & \textbf{WikiTQ} & \textbf{TabFact} & \textbf{HiTab} & \textbf{AiT-QA} \\
        \midrule
         Base Model             & 63.5                      & 76.0                          & 62.7                      & 55.5  \\
         + LTI (w/o rewriting)  & 59.2 ($\downarrow$ 4.3)   & 71.4 ($\downarrow$ 4.6)       & 58.2 ($\downarrow$ 4.5)   & 50.1 ($\downarrow$ 5.4) \\
         + LTI                  & 65.8 ($\uparrow  $ 2.3)   & 77.5 ($\uparrow  $ 1.5)       & 65.7 ($\uparrow  $ 3.0)   & 59.1 ($\uparrow  $ 3.6) \\
        \bottomrule 
    \end{tabular}
  }
  \caption{Impact of LTI on Qwen-7B across four TR benchmarks. Label rewriting is crucial—without it, performance drops. Full LTI yields consistent gains across all datasets.}
  \label{table: ablation2}
  \vspace{-1em} 
\end{wraptable}

We conduct ablation studies (RQ2) on four datasets based on two base models to assess each component of \mymodel{}. As shown in Table~\ref{table: ablation}, all modules enhance overall performance. Key findings include: (1) Cold start provides the largest gains, especially for weaker base models; (2) GRPO consistently boosts performance, with or without cold start; (3) Mix-GRPO yields notable improvements on structurally complex datasets (HiTab, AIT-QA).

To examine mix-paradigm GRPO (RQ2), we analyze output template distributions and observe a strong preference for P-TR (WTQ: 95\%, HiTab: 91\%), supporting our hypothesis that P-TR better suits SLMs.
Case studies further show that, with mix-paradigm enabled, the model selectively switches to text-based templates when contextually appropriate (The strategy is learned during training and tends to switch to T-TR when headers are dense and the question is a simple look-up. See Fig.~\ref{fig: case5}).

Table~\ref{table: ablation2} further investigates the effectiveness of the original SSL task—LTI—and the necessity of incorporating continual learning. 
As shown, using only template-based annotation (w/o rewriting) leads to catastrophic forgetting. In contrast, introducing continual learning significantly improves model performance.

\begin{figure}[!t]
    \centering
    \includegraphics[width=0.97\linewidth]{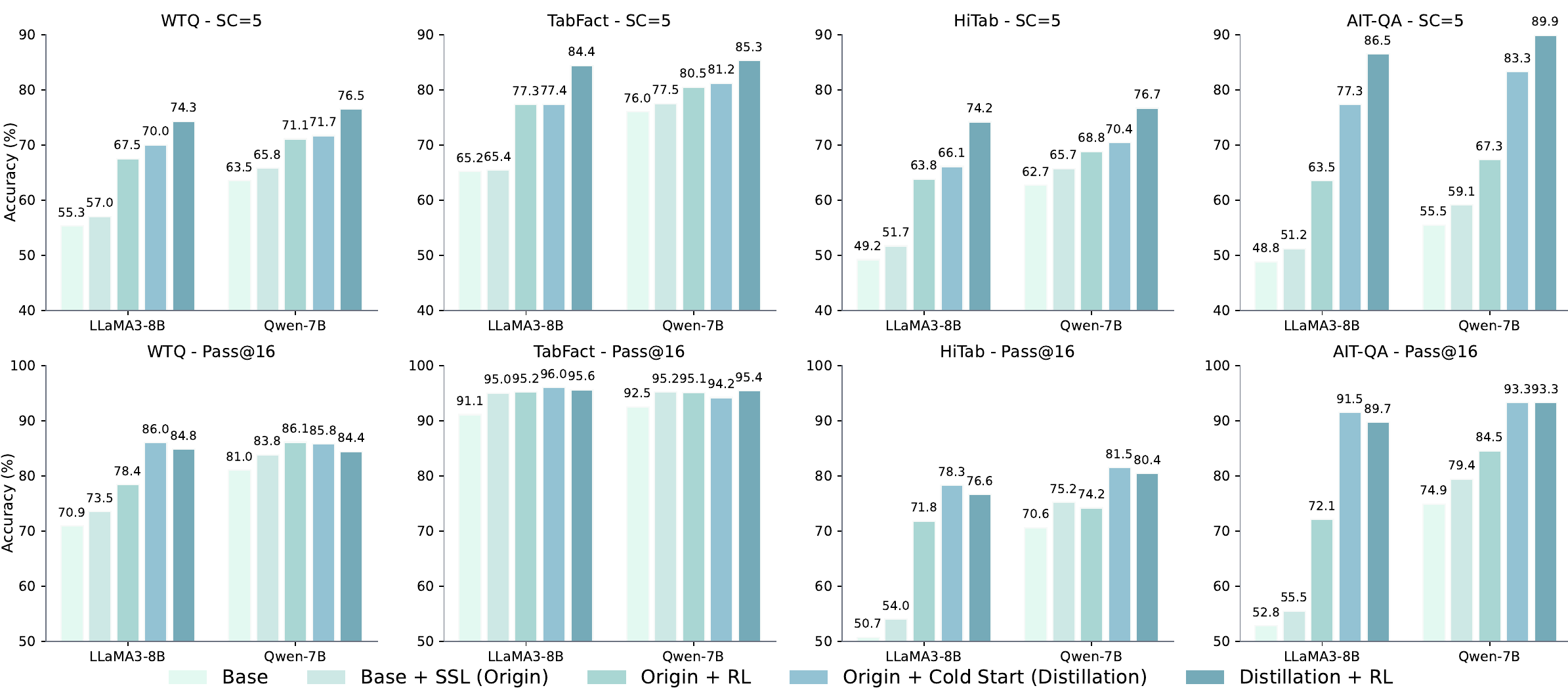}
    \caption{Comparison of SC=5 and Pass@16 accuracy across five model variants to assess the impact of SFT and RL. The results highlight the effects of SSL, distillation, and GRPO on model potential and capability.}
    \label{fig: results_2_4}
    \vspace{-0.3cm}
\end{figure}

To analyze the impact of RL and SFT on model performance (RQ3), we compare SC=5 and Pass@16 accuracy across five variants: Base, Origin (Base + SSL), Origin + RL, Distillation (Origin + Cold Start), and Distillation + RL. Results in Fig.~\ref{fig: results_2_4} yield three key observations:
TabFact exhibits minimal change in potential due to its binary nature, where random sampling yields a high probability of correct answers.
All components—SSL, distillation, and GRPO—improve both potential (Pass@16) and capability (SC=5), with SFT showing the most substantial gains.
While distillation + GRPO achieves the highest capability, GRPO after distillation reduces potential. Moreover, GRPO still cannot fully convert potential into actual capability.

\begin{figure}[!t]
    \centering
    \includegraphics[width=.9\linewidth]{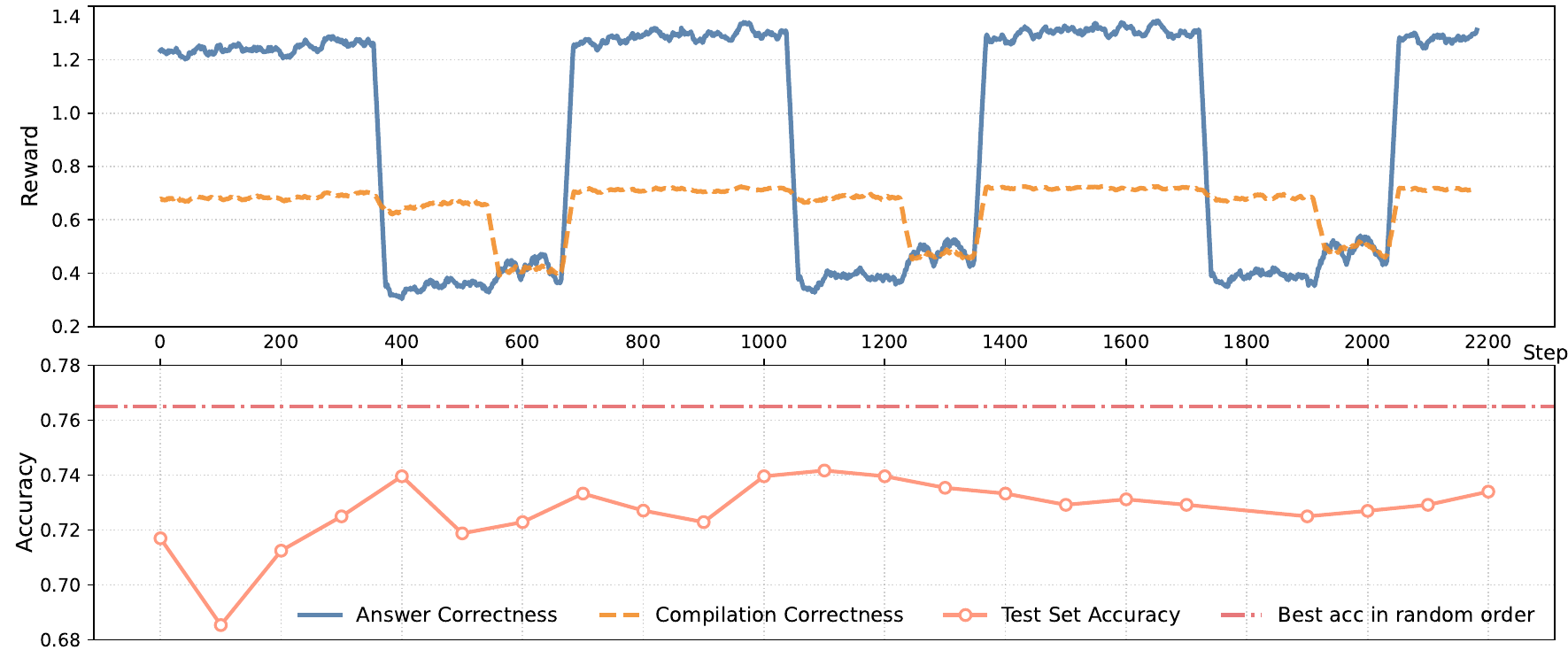}
    \caption{Training curves of Qwen-7B (after distillation) on the WTQ dataset sorted by difficulty levels (Easy → Medium → Hard) for 2200 steps. The plot shows reward signals during training and evaluation accuracy on the dev set. The red dashed line indicates the best accuracy achieved on WTQ with randomly shuffled training data.}
    \label{fig: level}
    \vspace{-0.2cm}  
\end{figure}

\subsection{Further Analysis (RQ4)}
Prior work~\citep{Understanding-r1-zero-like, zeng2025simplerl} has shown that GRPO is sensitive to training data difficulty. To further investigate this (RQ4), we conduct controlled experiments on WTQ, a benchmark with substantial P-TR challenges. Using execution feedback from a distilled Qwen-7B model, we categorize training examples into three levels:
Easy: code compiles and yields correct answers (7.5k examples);
Medium: code compiles but yields incorrect answers (3.5k);
Hard: code fails to compile (2.3k).
Each subset is trained separately for 2200 steps (20 instances per step), logging two reward signals: answer correctness (max 1.5) and compilation success (max 0.75). Test set accuracy is recorded every 100 steps.

Results (Fig.~\ref{fig: level}) show that compilation reward degrades with increasing difficulty, while answer correctness drops sharply even at the medium level. Throughout the process, models trained on sorted subsets underperform those trained on randomly shuffled data (76.5\% dev accuracy).
We also tested training only on the Easy+Medium subsets, achieving a best result of 75.5\%.
These findings suggest that while random mixing yields the best overall performance, training on data closer to the model's current ability can approach optimal results when compute resources are limited.

\section{Conclusion}
We present \mymodel{}, a two-stage method that enhances SLM-based TR through an original SSL task and mix-paradigm GRPO. It effectively mitigates layout generalization and reasoning inconsistency issues in P-TR, achieving performance competitive with LLMs.
Future work will explore broader RL strategies to advance TR capabilities further.


\bibliographystyle{iclrbib}
\bibliography{main}


\newpage
\appendix

\section{Limitations}
\label{appx: limitations}

1. Although our method has been evaluated on four popular public benchmarks—covering various types of semi-structured tables (e.g., WTQ, TabFact)—it may not generalize well to other types of tables, such as multimodal tables.

2. The SLMs used in this study refer to 7B or 8B-scale language models, and the proposed approach may not be directly applicable to smaller models such as 3B or 0.5B LMs.

3. Currently, the reward weighting for different functions in reinforcement learning still resembles neural network hyperparameters, heavily reliant on empirical tuning and lacking automation.

\section{Distinction Between Table Reasoning and Text-to-SQL}
While TR and Text-to-SQL (T2SQL) share the overarching goal of enabling LLMs to extract question-relevant information from tabular data, there remain fundamental differences between the two tasks—especially in light of recent work where symbolic approaches for TR also adopt SQL-style intermediate programs.

\textbf{Data Assumptions and Table Regularity.}
T2SQL operates over well-structured relational databases, where tables conform to strict schema constraints: each column has a clearly defined type, keys and relationships are explicitly annotated, and the table format is normalized. In contrast, tables in table reasoning tasks—such as those found in enterprise spreadsheets or scraped HTML tables—are typically semi-structured: they may lack consistent formatting, contain noisy or nested entries, and often do not enforce column type constraints (e.g., a single column might contain both strings and floats).

\textbf{Input Format and Model Access.}
Because T2SQL assumes clean schema-driven tables, models are typically given only the table schema (i.e., column headers and possibly data types or foreign key links), and must generate SQL queries accordingly. In Table Reasoning, however, the semi-structured and ad hoc nature of tables necessitates full-table access, requiring the model to parse and reason over raw cell content, layout artifacts, and potential multi-row/column dependencies.

\section{Implementation Details}
\label{appx: Implementation}
Our base models are Qwen2.5-Coder-7B and LLaMA3.1-8B-Instruct, with DeepSeek-v3 serving as the teacher model for the cold-start phase. For each dataset, we distill up to 5,000 samples. Both supervised fine-tuning (SFT) and reinforcement learning (RL) are implemented using LoRA, with a rank of 32 for SFT and 64 for RL. During RL, the decoding temperature is set to 0.85.

All experiments are conducted on NVIDIA RTX 4090 GPUs and NVIDIA H100 GPUs. The batch size is set to 4 × 8 × 8, where 4 is the number of samples per batch, the first 8 corresponds to the number of responses generated per sample, and the second 8 is the number of gradient accumulation steps. We train for 1 epoch.
We use the trl library for GRPO implementation and VLLM as the inference backend. The maximum token length is set to 2,400, and input tables are truncated to the first 10 rows. The model achieves its best performance at around 400 steps.

During the cold-start phase, the teacher model used is DeepSeek-v3, and the template employed for distillation is the P-TR template. The T-TR template was intentionally excluded because its inclusion was observed to degrade post-distillation performance.

\section{Predefined Operations for LTI and Reward Functions}
\label{appx: operations}
\FloatBarrier
\begin{table}[!h]
\centering
\begin{tabularx}{\linewidth}{XX}
\hline
Operations & Notes  \\
\hline
\texttt{row\_swap()}                                            & Swap the positions of two rows..\\
\texttt{column\_swap()}                                         & Swap the positions of two columns. \\
\texttt{row\_deletion()}                                        & Remove one or more rows.     \\
\texttt{column\_deletion()}                                     & Remove one or more columns.    \\
\texttt{extract\_rows()}                                        & Extract a subset of rows.    \\
\texttt{extract\_columns()}                                     & Extract a subset of columns.   \\
\texttt{extract\_rows\_having\_cells()}                         & extract rows containing cells with specific values.    \\
\texttt{extract\_columns\_having\_cells()}                      & extract columns containing certain cell values.   \\
\texttt{transpose()}                                            & Transpose the table, i.e., the top header becomes the left header, and vice versa.   \\
\hline
\end{tabularx}
\caption{Pre-defined Operations for LTI}
\label{table: Pre-defined_Operations}
\end{table}

\begin{table}[!h]
\centering
\begin{tabularx}{\linewidth}{XcX}
\hline
Reward Functions & Max Reward Value & Notes  \\
\hline
\texttt{strict\_format\_reward()}                       & 0.75    & Verifies that the entire output is strictly wrapped by tags. \\
\texttt{ans\_reward()}                                  & 1.5    & Measures the similarity between the model output and the reference answer. \\
\texttt{comment\_ratio}                                 & 0.45    & Evaluates the proportion of comments in the generated Python code. \\
\texttt{multiple\_python\_blocks\_}\\\texttt{penalty()} & 1.0    & Penalizes overly fragmented Python code with multiple separate blocks. \\
\texttt{compilation\_correctness}                       & 0.75   & Checks whether the generated code compiles without errors. \\
\texttt{code\_short\_length\_penalty()}                 & 0.5    & Penalizes code that is shorter than a defined length threshold. \\
\hline
\end{tabularx}
\caption{Reward Functions}
\label{table: reward_functions}
\end{table}

\FloatBarrier
\section{Prompts}
\label{appx: prompt}

\subsection{Prompts for T-TR}
\label{appx: prompt_t}
\begin{figure}[!h]
    \centering
    \includegraphics[width =1.0\linewidth]{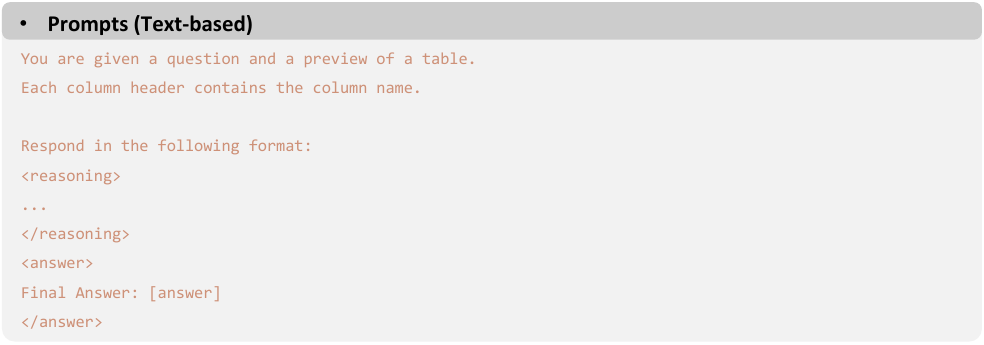}
    \caption{Prompts for T-TR.}
    \label{fig: prompts0}
\end{figure}

\subsection{Prompts for P-TR}
\label{appx: prompt_p}
\begin{figure}[!h]
    \centering
    \includegraphics[width =1.0\linewidth]{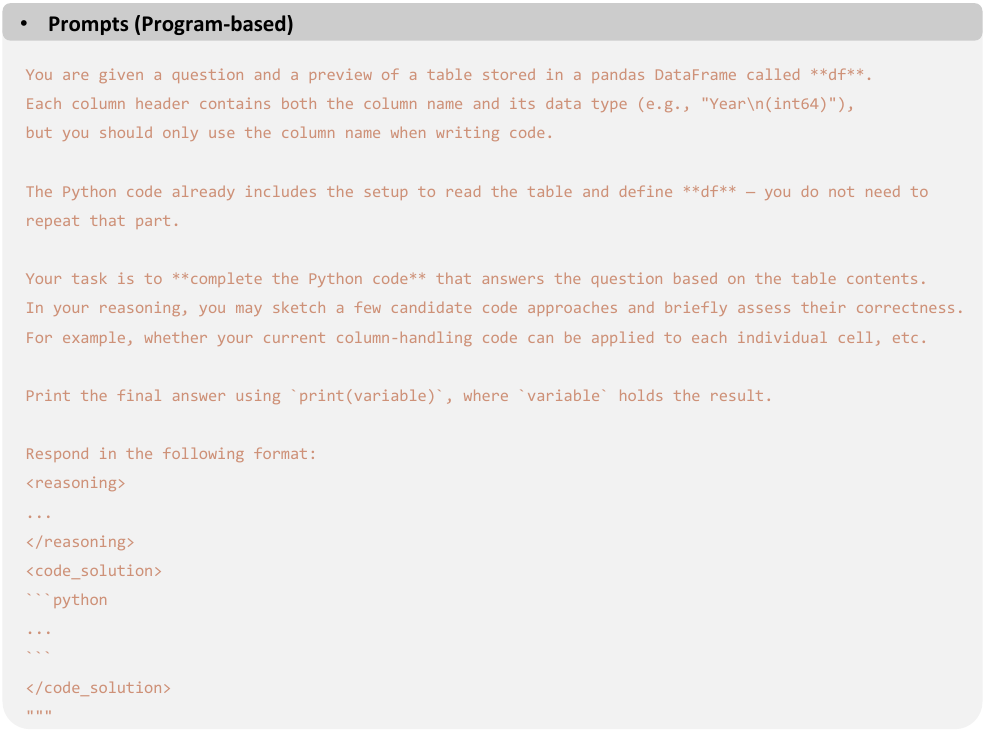}
    \caption{Prompts for P-TR.}
    \label{fig: prompts1}
\end{figure}

\subsection{Prompts for the Mix-paradigm GRPO}
\label{appx: prompt_grpo}
\begin{figure}[!h]
    \centering
    \includegraphics[width =1.0\linewidth]{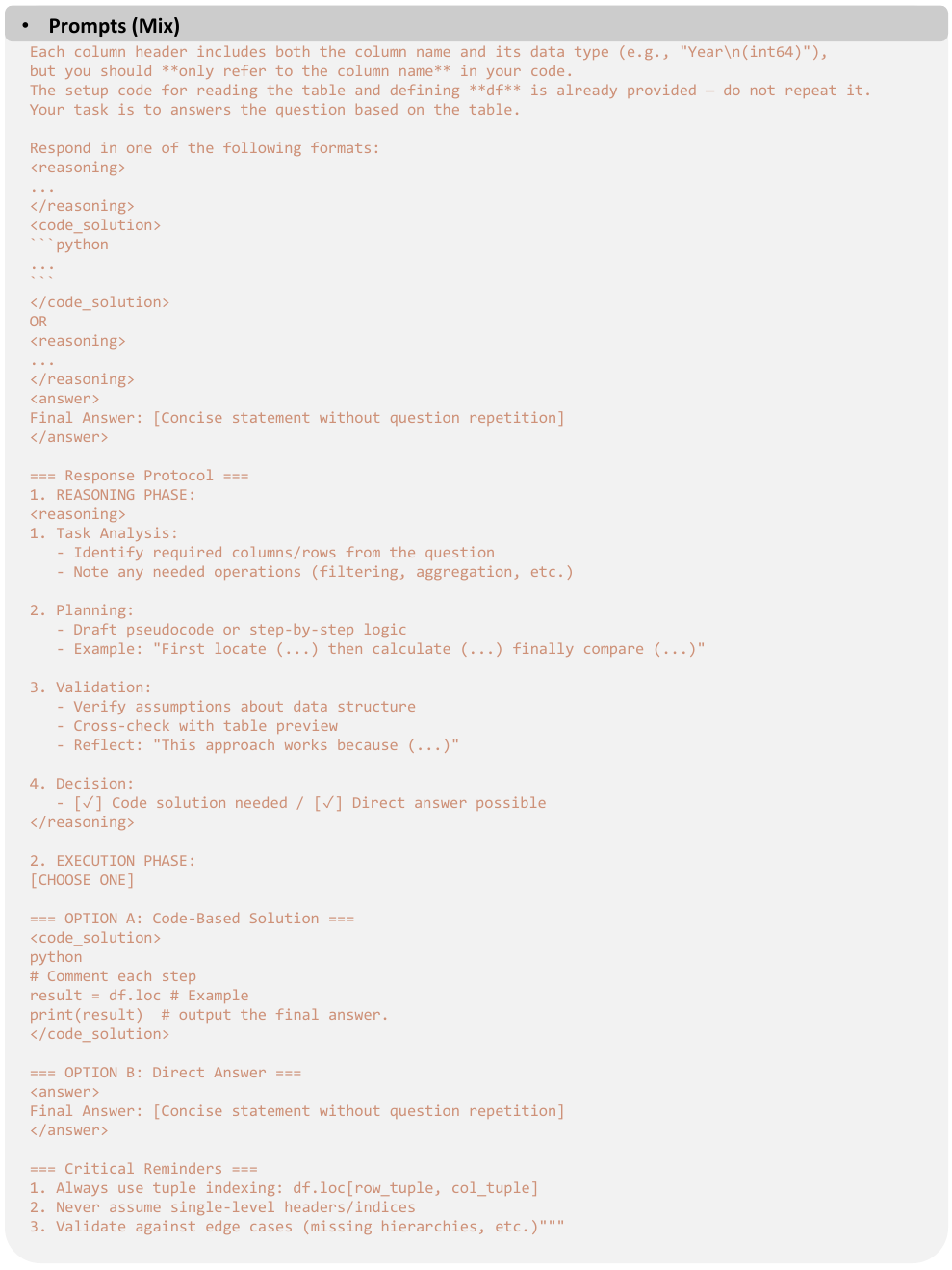}
    \caption{Prompts for the Mix-paradigm GRPO.}
    \label{fig: prompts2}
\end{figure}

\FloatBarrier

\section{Case Studies}
\label{appx: case}

\begin{figure}[!h]
    \centering
    \includegraphics[width =.85\linewidth]{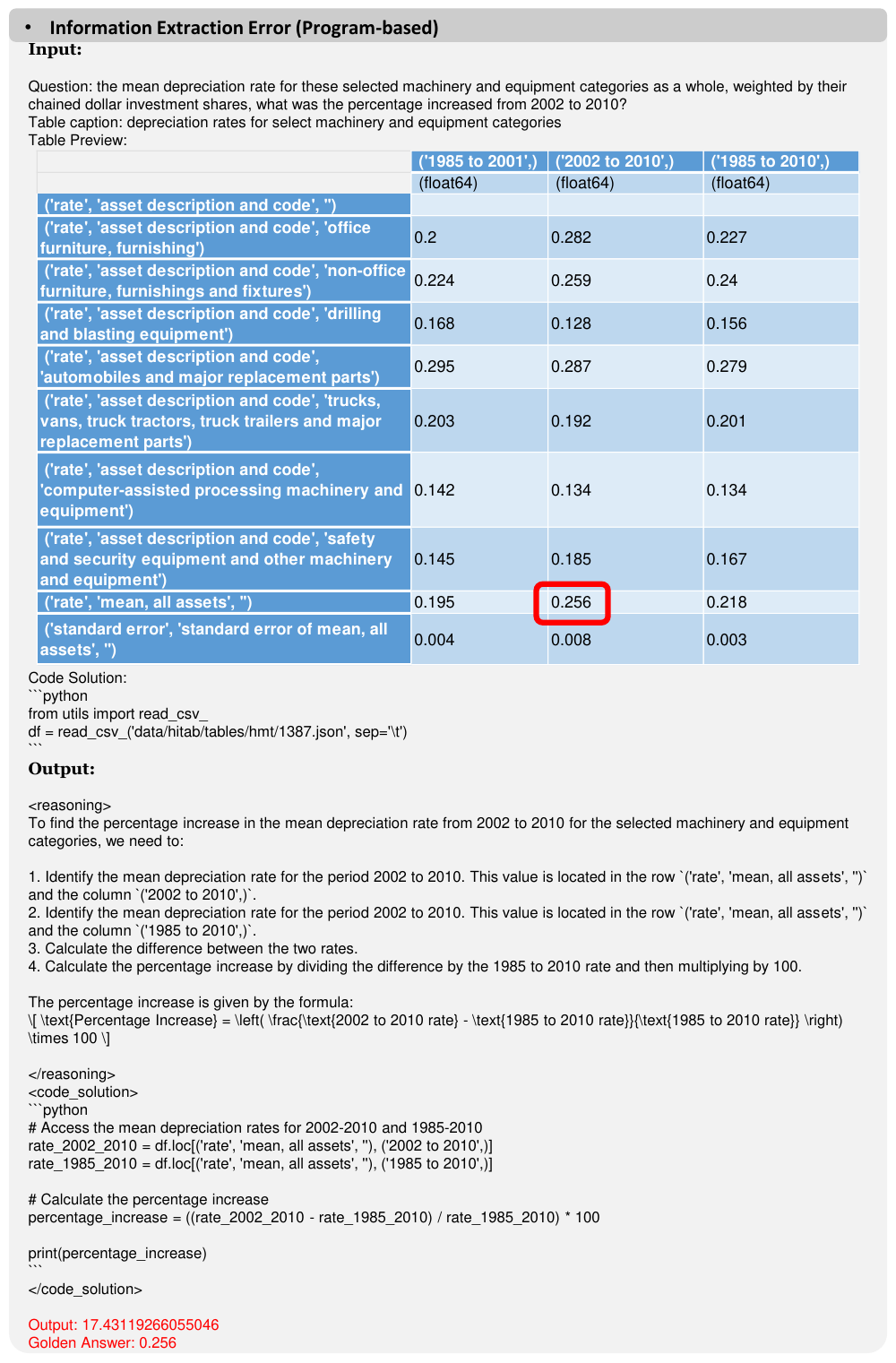}
    \caption{In P-TR, errors occur due to misunderstandings of the table's semantics. Specifically, the second-to-last row and second-to-last column of the table already provide the answer to the question, but the model fails to recognize this and instead incorrectly attempts to calculate the average from the existing numerical values.}
    \label{fig: case1}
\end{figure}

\begin{figure}[!h]
    \centering
    \includegraphics[width =1.0\linewidth]{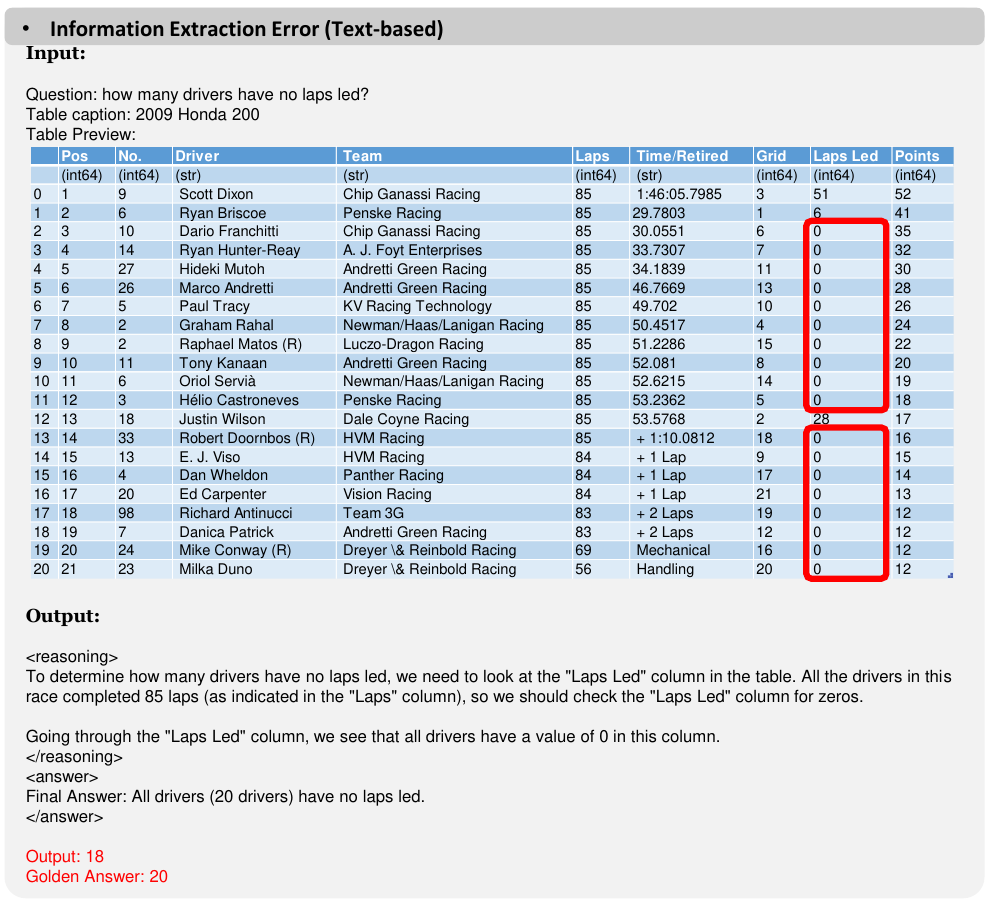}
    \caption{In T-TR, errors arise from information extraction errors of the table. Specifically, the second-to-last column contains non-zero cells, yet the model incorrectly assumes all cells to be zero. This also demonstrates that as table size increases, the model's ability to comprehend tabular information becomes progressively more challenging.}
    \label{fig: case2}
\end{figure}

\begin{figure}[!t]
    \centering
    \includegraphics[width =1.0\linewidth]{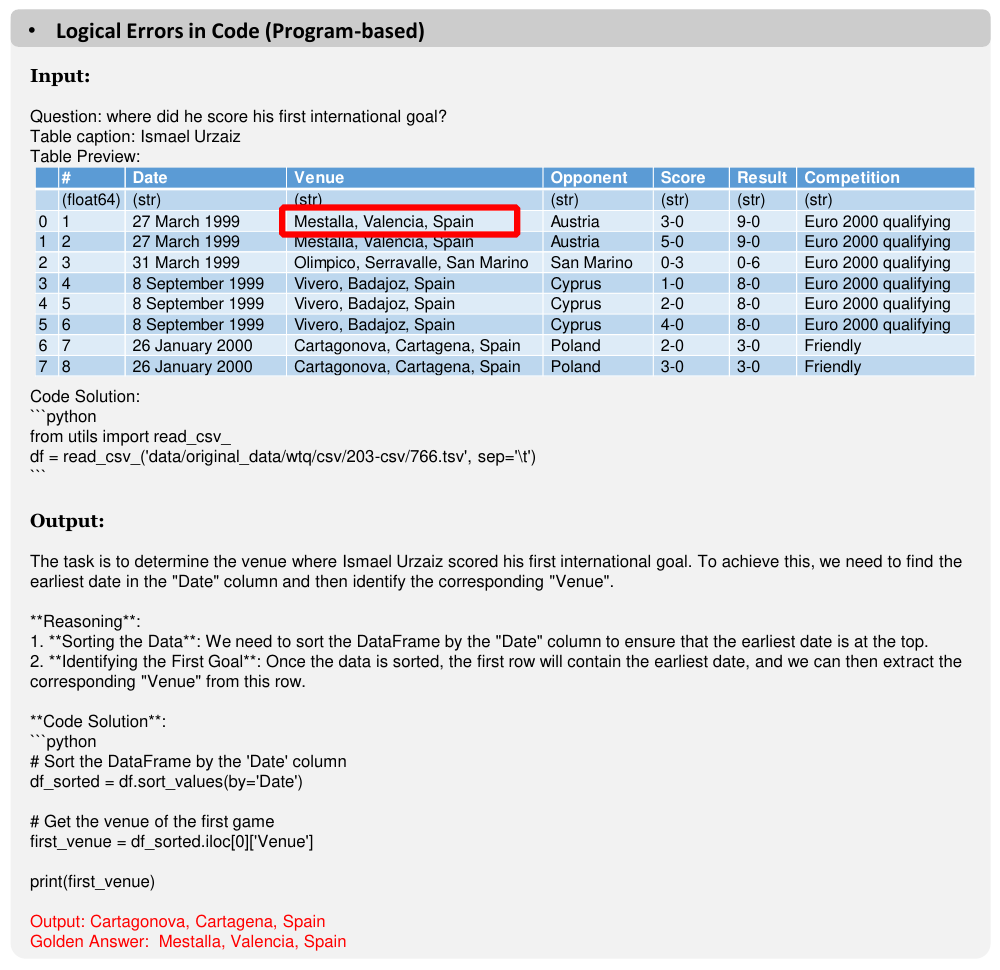}
    \caption{In P-TR, errors occur when the generated code solution lacks thorough consideration. Specifically, while the model is tasked with identifying the earliest timestamp, the implemented code merely performs a simple sort on the "Date" column without first converting the values into comparable numerical date formats.}
    \label{fig: case3}
\end{figure}

\begin{figure}[!t]
    \centering
    \includegraphics[width =1.0\linewidth]{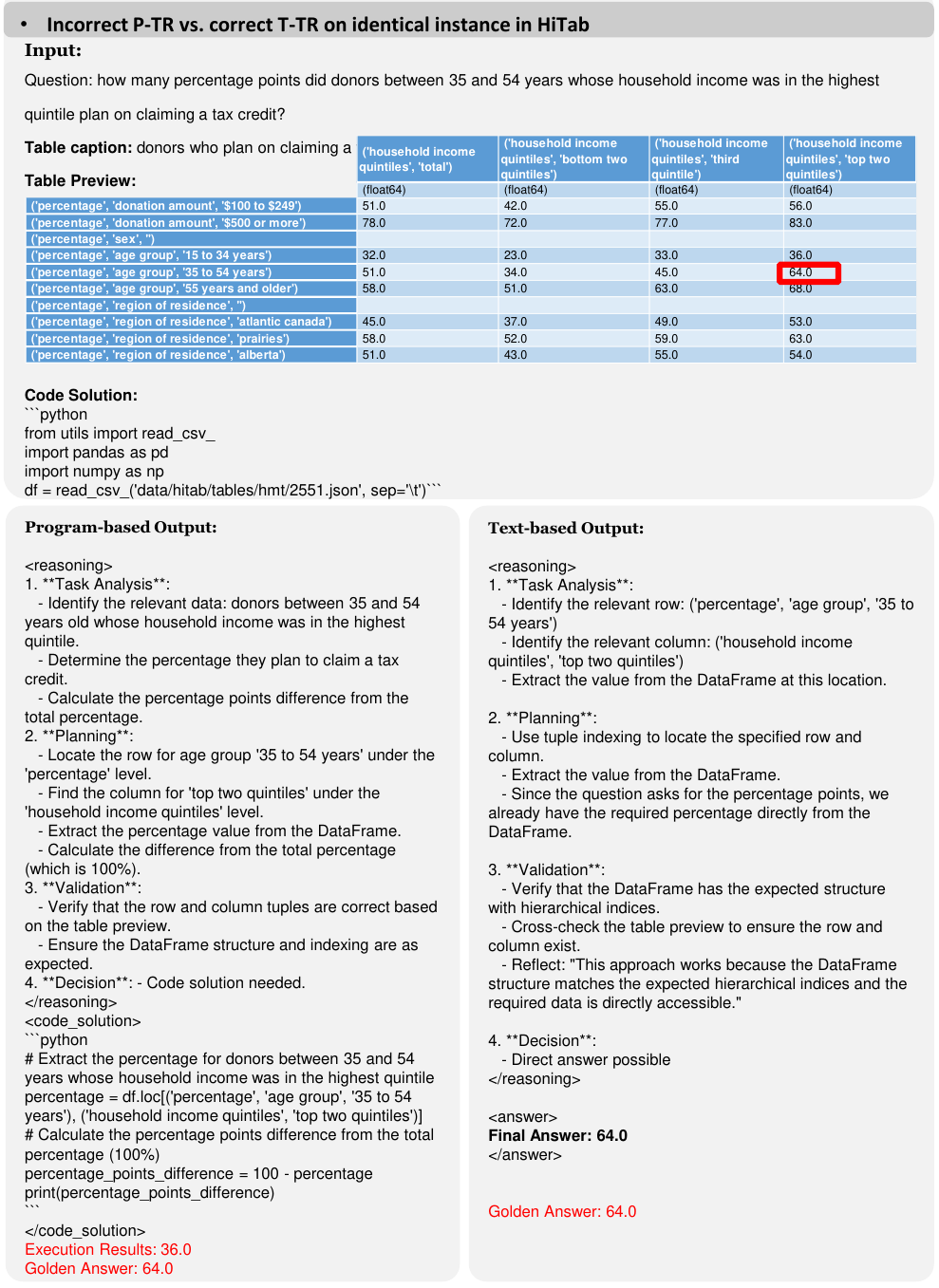}
    \caption{An example where the P-TR method fails but the T-TR method succeeds, due to the table layout being too complex for the program to correctly use headers as indices.}
    \label{fig: case5}
\end{figure}

\clearpage
\section*{NeurIPS Paper Checklist}

\begin{enumerate}

\item {\bf Claims}
    \item[] Question: Do the main claims made in the abstract and introduction accurately reflect the paper's contributions and scope?
    \item[] Answer: \answerYes{} 
    \item[] Justification: See Abstract and Section~\ref{section: intro}.

\item {\bf Limitations}
    \item[] Question: Does the paper discuss the limitations of the work performed by the authors?
    \item[] Answer: \answerYes{} 
    \item[] Justification: See Appx.~\ref{appx: limitations}.

\item {\bf Theory assumptions and proofs}
    \item[] Question: For each theoretical result, does the paper provide the full set of assumptions and a complete (and correct) proof?
    \item[] Answer: \answerNA{} 
    \item[] Justification: The paper does not include theoretical results.

    \item {\bf Experimental result reproducibility}
    \item[] Question: Does the paper fully disclose all the information needed to reproduce the main experimental results of the paper to the extent that it affects the main claims and/or conclusions of the paper (regardless of whether the code and data are provided or not)?
    \item[] Answer: \answerYes{} 
    \item[] Justification: See Section~\ref{section: settings} and Appx.~\ref{appx: Implementation}.

\item {\bf Open access to data and code}
    \item[] Question: Does the paper provide open access to the data and code, with sufficient instructions to faithfully reproduce the main experimental results, as described in supplemental material?
    \item[] Answer: \answerYes{} 
    \item[] Justification: We provide code along with sufficient instructions in supplemental material.

\item {\bf Experimental setting/details}
    \item[] Question: Does the paper specify all the training and test details (e.g., data splits, hyperparameters, how they were chosen, type of optimizer, etc.) necessary to understand the results?
    \item[] Answer: \answerYes{} 
    \item[] Justification: See Section~\ref{section: settings}, Appx.~\ref{appx: Implementation} and supplemental material.

\item {\bf Experiment statistical significance}
    \item[] Question: Does the paper report error bars suitably and correctly defined or other appropriate information about the statistical significance of the experiments?
    \item[] Answer: \answerYes{} 
    \item[] Justification: Standard deviation in Table~\ref{table: main-results}.

\item {\bf Experiments compute resources}
    \item[] Question: For each experiment, does the paper provide sufficient information on the computer resources (type of compute workers, memory, time of execution) needed to reproduce the experiments?
    \item[] Answer: \answerYes{} 
    \item[] Justification: See Appx.~\ref{appx: Implementation}.
    
\item {\bf Code of ethics}
    \item[] Question: Does the research conducted in the paper conform, in every respect, with the NeurIPS Code of Ethics \url{https://neurips.cc/public/EthicsGuidelines}?
    \item[] Answer: \answerYes{} 
    \item[] Justification: The research conducted in the paper conform, in every respect, with the NeurIPS Code of Ethics.

\item {\bf Broader impacts}
    \item[] Question: Does the paper discuss both potential positive societal impacts and negative societal impacts of the work performed?
    \item[] Answer: \answerNA{} 
    \item[] Justification: There is no societal impact of the work performed.
    
\item {\bf Safeguards}
    \item[] Question: Does the paper describe safeguards that have been put in place for responsible release of data or models that have a high risk for misuse (e.g., pretrained language models, image generators, or scraped datasets)?
    \item[] Answer: \answerNA{} 
    \item[] Justification: The paper poses no such risks.

\item {\bf Licenses for existing assets}
    \item[] Question: Are the creators or original owners of assets (e.g., code, data, models), used in the paper, properly credited and are the license and terms of use explicitly mentioned and properly respected?
    \item[] Answer: \answerYes{} 
    \item[] Justification: See Section~\ref{section: settings}.

\item {\bf New assets}
    \item[] Question: Are new assets introduced in the paper well documented and is the documentation provided alongside the assets?
    \item[] Answer: \answerNA{} 
    \item[] Justification: The paper does not release new assets.

\item {\bf Crowdsourcing and research with human subjects}
    \item[] Question: For crowdsourcing experiments and research with human subjects, does the paper include the full text of instructions given to participants and screenshots, if applicable, as well as details about compensation (if any)? 
    \item[] Answer: \answerNA{} 
    \item[] Justification: The paper does not involve crowdsourcing nor research with human subjects.

\item {\bf Institutional review board (IRB) approvals or equivalent for research with human subjects}
    \item[] Question: Does the paper describe potential risks incurred by study participants, whether such risks were disclosed to the subjects, and whether Institutional Review Board (IRB) approvals (or an equivalent approval/review based on the requirements of your country or institution) were obtained?
    \item[] Answer: \answerNA{} 
    \item[] Justification: The paper does not involve crowdsourcing nor research with human subjects.

\item {\bf Declaration of LLM usage}
    \item[] Question: Does the paper describe the usage of LLMs if it is an important, original, or non-standard component of the core methods in this research? Note that if the LLM is used only for writing, editing, or formatting purposes and does not impact the core methodology, scientific rigorousness, or originality of the research, declaration is not required.
    \item[] Answer: \answerYes{} 
    \item[] Justification: LLMs for reasoning and distillation.

\end{enumerate}

\end{document}